\pdfoutput=1

\documentclass[11pt]{article}

\usepackage[]{acl}

\usepackage{times}
\usepackage{latexsym}
\usepackage{graphicx}
 \usepackage{booktabs}
\usepackage[T1]{fontenc}

\usepackage[utf8]{inputenc}

\usepackage{microtype}

\title{Discovering Customer-Service Dialog System with Semi-Supervised Learning and Coarse-to-Fine Intent Detection}
\author{Zhitong Yang, Xing Ma, Anqi Liu, Zheyu Zhang, Chuankai Xu, Hanqing Hu, \\ 
{\bf Wenjie Bai, Chen Wang, Yihong Tang, Yipeng Xu, Shangjing Huang,} \\
 {\bf Dongming Zhao, Peng Zhang and Bo Wang  } \\ 
    School of New Media and Communication, Tianjin University, Tianjin, China \\
    College of Intelligence and Computing, Tianjin University, Tianjin, China \\
    AI Lab, China Mobile Communication Group Tianjin Co., Ltd.\\
    \texttt{\{yyyyyyzt, machine981\}@tju.edu.cn}}

\begin{document}
\maketitle
\begin{abstract}
Task-oriented dialog(TOD) aims to assist users in achieving specific goals through multi-turn conversation. Recently, good results have been obtained based on large pre-trained models. However, the labeled-data scarcity hinders the efficient development of TOD systems at scale. In this work, we constructed a weakly supervised dataset based on a teacher/student paradigm that leverages a large collection of unlabelled dialogues. Furthermore, we built a modular dialogue system and integrated coarse-to-fine grained classification for user intent detection. Experiments show that our method can reach the dialog goal with a higher success rate and generate more coherent responses.
\end{abstract}
\begin{figure*}[htbp]
\includegraphics[width=1\textwidth]{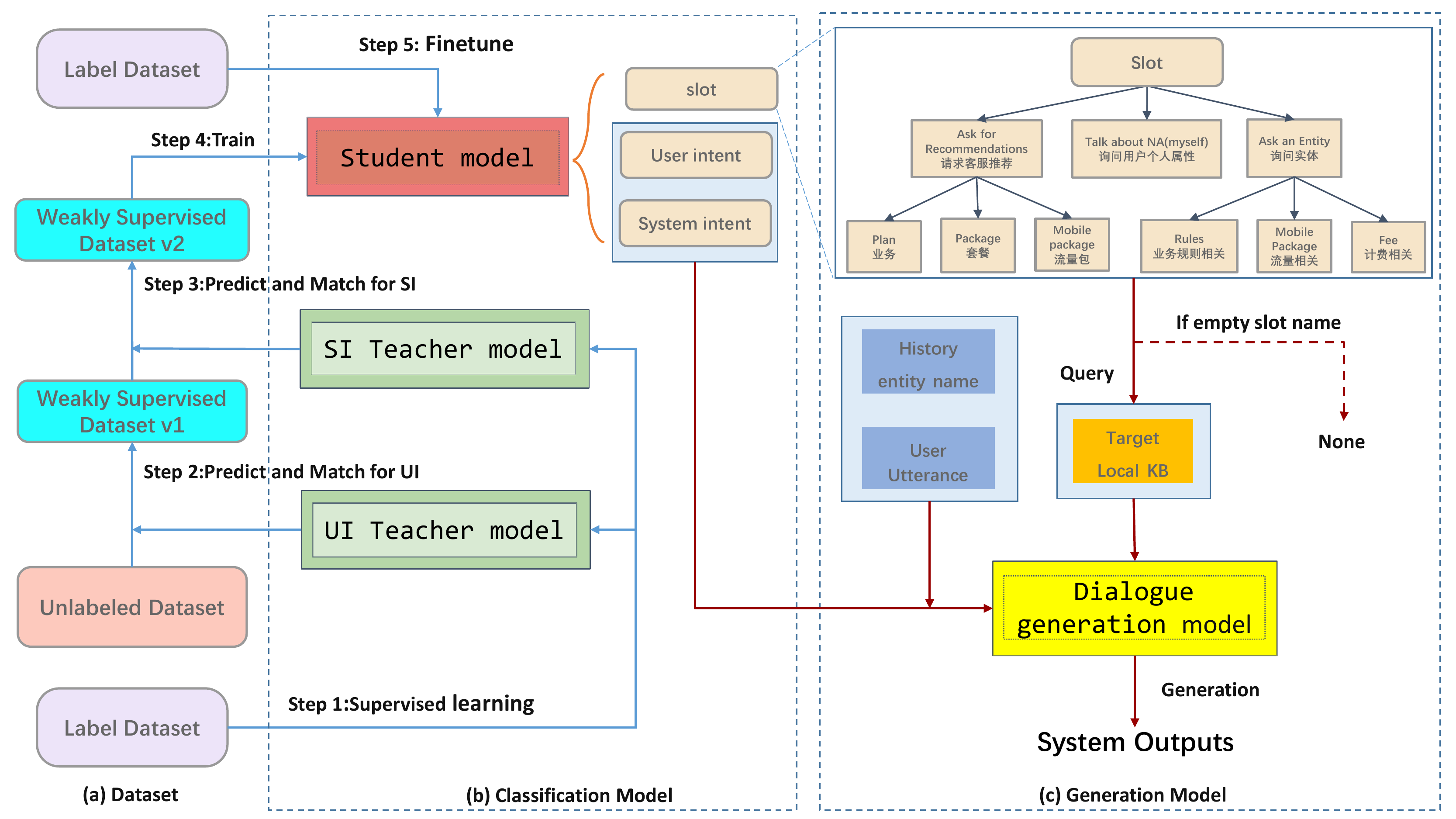}
\caption{Illustration of our model architecture. (a)The semi-supervised learning pipeline to train the student model. (b) Our coarse-to-fine user intent classification model which described in section \ref{ctf}. (c) Our generation model takes user intent(UI), service intent(SI), history entity name, user utterance, and target local KB as input.}
\label{fig:overall}
\end{figure*}
\section{Introduction}

Current task-oriented dialogue systems are often modeled by pipeline \citep{young2013pomdp,mrkvsic2016neural} or end-to-end methods \citep{wen2016network,lei2018sequicity, madotto2018mem2seq,wu2019global,qin2020dynamic}. Traditional pipeline methods divide it into four subtasks including: 1) Dialog understanding; 2) Dialog state tracking; 3) Dialog policy planning; 4) Dialog Generation. It is more controllable but causes error propagation problems probably. Recently, good results have been obtained using more general architectures based on end-to-end models. They are easy to train and effectively alleviates the problem of error propagation but lacks some interpretability. However, training TOD systems requires manually labeled dialog states and system acts, which is costly to obtain. 

Remarkably, unlabeled data are often readily available in many forms, such as human-to-human dialogs. This has motivated the development of semi-supervised learning (SSL) \citep{zhu2005semi}, which aims to leverage both labeled and unlabeled data. \citet{lee2013pseudo} used the model trained on labeled data to generate labels for unlabeled data. \citet{verma2019interpolation} improved classification performance by minimizing the discrepancy between two probability distributions by perturbed unlabeled data. \citet{wu2021r} proposed a similar method, R-drop, to reduce the divergence of the output distributions between two sub-models sampled by dropout. \citet{gao2021simcse} also leveraged dropout to construct positive instances and used a contrastive objective to learn sentence representations. However, the effect of labeled data on neural models is more significant. Especially when the labeled data presents a long-tailed distribution or has class-imbalance problems, the effect of semi-supervised methods is limited. Implementing an effective semi-supervised method is challenging when the quality of labeled data is poor.
 
In this paper, we use semi-supervised learning to fully utilize unlabeled conversations to build a customer-service task-oriented dialogue system and incorporate coarse-to-fine grained intent detection for more accurate KB queries. Specifically, a new weakly training set is constructed from large-scale unlabeled data through a teacher-student framework. We train two well-performed teacher models to predict labels for unlabeled data. At the same time, we cluster the user's intent to query KB and obtain three coarse-grained categories. Based on this, a student model with coarse-to-grained intent detection is trained on the weakly supervised dataset and fine-tuned on the original dataset. Finally, we use a pre-trained model to combine the query results to give a response. We build our dialogue system and conduct experiments on the MobileCS dataset with 10,000 labeled data and 90,000 unlabeled data. Our system achieves excellent results on the dataset and outperforms baseline models on all metrics.

In summary, our main contributions include three aspects:
\begin{itemize}
\item[1)]
Our system makes full use of unlabeled data to provide a weakly supervised dataset. In contrast to most semi-supervised methods, we set up two teacher models for teaching according to user roles(user and customer service).
\item[2)]
We design a coarse-to-fine grained intent detection model and a knowledge-augmented response generation model to build the TOD systems. Specifically, we divide the coarse-to-fine intent into a tree structure, as shown in figure \ref{fig:overall}.
\item[3)]
We conduct sufficient experiments to verify that our model outperforms well on MobileCS dataset and win the top-3 in SereTOD Challenge. 
\end{itemize}

\section{Our Approach}

\subsection{Method Overview}

Our System consists of three main parts: (1) Weakly supervised data extracted by semi-supervised learning, (2) a coarse-to-fine grained intent classification model is used to predict user intent, slot, and service intent, and (3) a controllable generative model for giving a response based on KB query results.

\subsection{Weakly supervised data extraction}

Based on the teacher-student framework\citep{yalniz2019billion}, our semi-supervised learning pipeline is as follows:

(1) Train a UI(user-intent) teacher model and a SI(service-intent) teacher model on labeled data $\mathcal{D}$, respectively;

(2) Run the trained UI model on unlabeled data $\mathcal{U}$ and select high-precision  examples to construct a new labeled dataset $\hat{\mathcal{D_\mathrm{u}}}$;

(3) Run the trained SI model on $\hat{\mathcal{D_\mathrm{u}}}$ and filter high precision examples to construct a new labeled dataset $\hat{\mathcal{D}}$;

(4) Train a new student model on $\hat{\mathcal{D}}$;

(5) Fine-tune the trained student on $\mathcal{D}$.

\noindent\textbf{Teacher model training} A good teacher model is required for this stage. It is responsible for removing unreliable examples from $\mathcal{U}$ to correctly label enough relevant examples without introducing substantial labeling noise. So we set up teacher models for the user and customer service roles. Note that we predict the SI(service-intent) in the original task based on the user input. But when we train the SI teacher model, we label the data according to the customer service response. 

\noindent\textbf{Data selection threshold} We set a probability threshold $\rho$ for each category to select the data. As expected, the probability threshold of introducing false positive becomes higher as $\rho$ decreases. Therefore, there is an important trade-off on $\rho$ and we set the threshold $\rho$ for each class according to its distribution.

\subsection{Coarse-to-Fine grained Intent Detection}\label{ctf}

\noindent\textbf{Pretraing} Firstly, we pre-trained BERT\citep{devlin2018bert} in the manner of unsupervised SimCSE\citep{gao2021simcse} with unlabeled data. Contrastive learning is applied in the process. In addition to dropout, we randomly replace some token in the sentence with the character `\_' for regularization. 

\noindent\textbf{Coarse-to-Fine} According to KB intent statistics for user queries, we divide the query slot of user-intent into three types: ``Talk\ about\ NA(myself)', ``Ask\ for\ Introduction', ``Ask\ an\ Entity'. As shown in the right part of figure \ref{fig:overall}, ``Ask\ for\ Introduction' can subdivided into (Plan, Package plan, Mobile package) and ``Ask\ an\ Entity' can be subdivided into (Rules, Mobile Package, Fee), respectively. After dividing the coarse-grained intentions in this way, the $Recall$ of predicting user intentions and  the $Success \ Rate$ will be improved.

\noindent\textbf{Joint Training} After redefining the classification method, we use three linear layers to connect BERT for joint training. To realize the multi-classification of intents in different classes and slots in coarse-to-fine categories, the BCE loss between the class probability and the correct label was used for the joint class prediction:

\begin{equation}\label{e1}
L = -\sum_{i=1}^{N} [y_{i}ln(\sigma (x_{i} ))+(1-y_{i} )ln(1-\sigma (x_{i} ))]
\end{equation}

where $\sigma (x )=\frac{1}{1+e^{-x} } $ is the predicted result, and $y_i$ is the real label. For a given dialogue, the total loss is a linear combination of $L_{ui}, L_{si}$ and $L_{slot}$:

\begin{equation}\label{e2}
loss=L_{ui}+L_{si}+L_{slot}
\end{equation}

At the same time, we set different weights for different categories so that the loss weights of each type are personalized.

\noindent\textbf{The intent `Other'} In addition, knowing that the data in the `other' category is difficult to distinguish and will seriously affect the experimental results, we do not explicitly predict the "Other" category of data. When predicting the intent, we classify data with a probability less than 0.1 of each class as `Other', which naturally matches the essential characteristics of the `other' category.

\subsection{Knowledge-augmented Response Generation}

We use pre-trained GPT-2 to generate a response and fine-tune it similarly to the baseline model. We convert raw data to sequences in a specific order, as shown in Figure \ref{fig:overall}. Then we use language modeling as a pre-training task and adopt cross-entropy loss.

In the inference stage, we give GPT-2 the history entities and user utterances as context to predict current entities auto-regressively. Next, the corresponding user and service intent classification results are added to the generated text. Then GPT-2 generates a system response based on the existing context.

In particular, when getting exact KB results from the classification module, our system encodes the KB results and forces GPT-2 to generate them in response through beam searching. Instead of predicted service intent, our system adds `inform' intention to the generated text to prevent simple responses and improve the dialog success rate.

\section{Experiments}

\subsection{Dataset}

\noindent\textbf{MobileCS dataset} MobileCS \citep{ou2022challenge} is a publicly available multi-domain TOD dataset and consists of real-world dialog data between users and customer-service staff from China Mobile. Dataset has 10,000 dialogs labeled by crowdsourcing, and the remaining 90,000 dialogs are unlabeled. The dataset contains many oral expressions and some inaccurate labels, making it more challenging. 

\noindent\textbf{Weakly supervised data} We use teacher-student schema to construct a new dataset from unlabeled data. We use two teacher models to make classification predictions on the unsupervised data and take the data with higher confidence as an additional training set. 300K dialog turns are acquired through this way. 

\begin{table*}[htbp]
 \small
 \centering
 \begin{tabular}{ccccccccc}\toprule
    & \multicolumn{3}{c}{\textbf{User intent}} & \multicolumn{3}{c}{\textbf{Service intent}} & \multicolumn{1}{c}{\textbf{BLEU-4}} & \multicolumn{1}{c}{\textbf{Success rate}} 
    \\\cmidrule(lr){2-4}\cmidrule(lr){5-7}
                        & Pre  & Rec & f1    & Pre  & Rec & f1\\\midrule
    \textbf{MGA-GPT}             & 0.677 & 0.626 & 0.651 & 0.616 & 0.637 & 0.574 & 4.17 & 0.315 \\
    \textbf{UniLM}               & 0.743 & 0.624 & 0.678 & 0.619 & 0.535 & 0.575 & 4.90 & 0.294\\
    \textbf{BERT+MemNN+GPT2}     & 0.775 & 0.609 & 0.682 & 0.587 & 0.609 & \textbf{0.598} & 6.49 & 0.550 \\
    \textbf{Our system}          & \textbf{0.807} & \textbf{0.647} & \textbf{0.718} & \textbf{0.606} & \textbf{0.590} & \textbf{0.598} & \textbf{7.54} & \textbf{0.689}
    \\\bottomrule
 \end{tabular}
 \caption{Evaluation results of our system and baselines on test data of MobileCS dataset. It shows precision, recall and f1 score of predicted user and service intent, success rate of dialog, and BLEU-4 score of responses.}
 \label{test result}
\end{table*}

\begin{table*}[htbp]
 \small
 \centering
 \begin{tabular}{lccccccccc}\toprule
    & \multicolumn{5}{c}{\textbf{Auto Evaluation}} & \multicolumn{4}{c}{\textbf{Human Evaluation}}
    \\\cmidrule(lr){2-6}\cmidrule(lr){7-10}
                                & \textbf{UI f1} & \textbf{SI f1} & \textbf{BLEU} & \textbf{Succ.} & \textbf{Comb.} & \textbf{Fluency} & \textbf{Coherency} & \textbf{Succ.} & \textbf{Aver.}
    \\\midrule
    \textbf{Team-2}             & 0.656 & 0.533 & 3.82 & 0.259 & 1.524 & 2.87 & 2.55 & 2.42 & 2.61\\
    \textbf{Team-5}             & \textbf{0.714} & \textbf{0.589} & 6.79 & 0.432 & 1.871 & 4.06 & \textbf{3.14} & \textbf{3.14} & \textbf{3.53}\\
    \textbf{Team-8}             & 0.699 & 0.550 & 6.44 & 0.644 & 2.022 & 2.39 & 2.29 & 2.03 & 2.24\\
    \textbf{Team-9}             & 0.682 & 0.587 & 3.63 & 0.217 & 1.458 & 3.20 & 2.98 & 3.11 & 3.10\\
    \textbf{Team-11}            & \textbf{0.728} & \textbf{0.595} & \textbf{14.43} & \textbf{0.780} & \textbf{2.392} & \textbf{4.23} & \textbf{3.73} & \textbf{3.47} & \textbf{3.81}\\
    \textbf{Our system}         & 0.706 & 0.587 & \textbf{7.76} & \textbf{0.669} & \textbf{2.117} & \textbf{3.55} & 3.03 & 2.77 & 3.12
    \\\bottomrule
 \end{tabular}
 \caption{Evaluation results (including auto metrics and human evaluation) of all teams in the SereTOD Challenge on final test data. Top-2 results for each metric are bold.}
 \label{eval result}
\end{table*}

\subsection{Baselines}

We compare our system with the following baseline models.
~\\

\noindent$\bullet$ \textbf{MGA-GPT} \citep{liu2022revisiting}: 
The model only takes current dialog turn into pre-trained model instead of the whole dialog history. Thourgh this way the efficiency of the model can be improved without too much impact on the performance of the model.
~\\

\noindent$\bullet$ \textbf{UniLM} \citep{dong2019unified}:
We follow the same way to GALAXY \cite{he2022galaxy} to pre-train and fine-tune UniLM. Then we use UniLM to generate intents and responses auto-gressively.
~\\

\noindent$\bullet$ \textbf{BERT+MemNN+GPT2} \citep{madotto2018mem2seq}:
We combine memory network with pointer and use copy mechanism to directly extract the entity attribute value or attribute name required for response generation from the KB instead of classifying slots.
~\\

We conduct experiments on MobileCS dataset with baseline models above and our system.

\subsection{Results\footnote{After the competition official released the labeled test set, we re-ran our programs and got the latest results}}

We adopt intent f1 score, success rate and BLEU-4 metrics to evaluate our model performance. The results on MobileCS dataset are shown in Table 1, we can observe that our system outperforms MGA-GPT and UniLM (generative model) by 37.4\% and 39.5\%. Although we combine MemNN with pointer to improve the performance of KB query, our system still get higher success rate than MemNN by 13.9\%, which indicates that our system is better at KB querying. Besides, our system and BERT+MemNN+GPT-2 get better intent f1 score which shows that BERT pre-trained with augmented data can improve the classifying performance effectively. 

We conduct experiments on final labeled test set released by official. Comparing with models of other teams, our system still has good performance especially on success rate and BLEU-4. We also get positive results in human evaluation which also proves the effectiveness of our system.

\section{Conclusion}

In this paper, we presented a modular task-oriented dialogue system with semi-supervised learning. We first mine a weakly supervised dataset from the MobileCS dataset based on the teacher-student paradigm. Then, we clustered the user's intent to query KB and obtained three coarse-grained categories. Based on this, a student model with coarse-to-fine grained intent detection is trained on the weakly dataset and fine-tuned on the original dataset. Automatic and human evaluations show that our method exceeds the baselines, and the weakly supervised dataset contributes significantly. We hope the proposed system in this work will be helpful for future work.

\label{sec:bibtex}


\bibliography{anthology,custom}
\bibliographystyle{acl_natbib}




\end{document}